\documentclass[letterpaper, 10 pt, conference]{ieeeconf}  

\IEEEoverridecommandlockouts                              

\usepackage{amsmath,amssymb,amsfonts}

\usepackage{amsthm}
\usepackage{subcaption}
\usepackage[english]{babel}
\usepackage{algorithmic}
\usepackage{algorithm}
\usepackage{graphicx}
\usepackage{textcomp}
\usepackage{xcolor}
\def\BibTeX{{\rm B\kern-.05em{\sc i\kern-.025em b}\kern-.08em
    T\kern-.1667em\lower.7ex\hbox{E}\kern-.125emX}}

\usepackage[font={sf,scriptsize},           
            caption=false]                  
           {subfig}

\usepackage[
    style=ieee
]{biblatex}

\usepackage{subfiles}
\usepackage{multirow}

\newcommand{\red}{\textcolor{red}}

\usepackage{url}
\usepackage{balance}
\usepackage[capitalize]{cleveref}

\title{\LARGE \bf
LiDAR Inertial Odometry And Mapping\\Using Learned Registration-Relevant Features
}

\author{Zihao Dong$^1$, Jeff Pflueger$^1$, Leonard Jung$^1$, David Thorne$^2$\\
Philip R. Osteen$^3$, Christa S. Robison$^3$, Brett T. Lopez$^2$, Michael Everett$^1$
\thanks{*This research was sponsored by the DEVCOM Army Research Laboratory (ARL) under SARA CRA
W911NF-24-2-0017. Distribution Statement A: Approved for public release; distribution is unlimited.}
\thanks{$^{1}$Northeastern University, Boston, MA, USA. \{\texttt{dong.zih, pflueger.j, jung.le, m.everett\}@northeastern.edu}}%
\thanks{$^{2}$University of California Los Angeles, Los Angeles, CA, USA. \texttt{\{davidthorne, btlopez\}@ucla.edu}}%
\thanks{$^{3}$DEVCOM Army Research Laboratory (ARL). \{\texttt{philip.r.osteen, christopher.j.robison5\}.civ@army.mil}}%
\thanks{Code: 
\protect\url{https://github.com/neu-autonomy/FeatureLIOM}
}
}

\begin{document}

\maketitle
\thispagestyle{empty}
\pagestyle{empty}

\begin{abstract}
SLAM is an important capability for many autonomous systems, and modern LiDAR-based methods offer promising performance.
However, for long duration missions, existing works that either operate directly the full pointclouds or on extracted features face key tradeoffs in accuracy and computational efficiency (e.g., memory consumption).
To address these issues, this paper presents DFLIOM with several key innovations.
Unlike previous methods that rely on handcrafted heuristics and hand-tuned parameters for feature extraction, we propose a learning-based approach that select points relevant to LiDAR SLAM pointcloud registration.
Furthermore, we extend our prior work DLIOM with the learned feature extractor and observe our method enables similar or even better localization performance using only about 20\% of the points in the dense point clouds. 
We demonstrate that DFLIOM performs well on multiple public benchmarks, achieving a 2.4\% decrease in localization error and 57.5\% decrease in memory usage compared to state-of-the-art methods (DLIOM).
Although extracting features with the proposed network requires extra time, it is offset by the faster processing time downstream, thus maintaining real-time performance using 20Hz LiDAR on our hardware setup.
The effectiveness of our learning-based feature extraction module is further demonstrated through comparison with several handcrafted feature extractors.
\end{abstract}

\section{Introduction}

Simultaneous Localization and Mapping (SLAM), the ability to provide accurate state estimates and maps, is crucial for robust and safe interaction with new environments \cite{palanisamy2020multi, huorbitgrasp, dong2024collision}.
This paper builds on LiDAR-based odometry methods~\cite{zhang2017low, shan2018lego, shan2020lio, xu2022fast, chen2023dlio, chen2023dliom}, which typically use point cloud registration algorithms, e.g., variants of iterative closest point (ICP)~\cite{besl1992method,segal2009generalized}, to estimate transformations between poses at different times.
Because LiDAR sensors can provide accurate, long-range depth measurements across a wide variety of environment conditions, today's LiDAR-based odometry methods are generally more accurate than vision-based methods~\cite{leutenegger2015keyframe, geneva2020openvins, huai2022robocentric}.
However, several research challenges remain for achieving robust and efficient LiDAR odometry.

\begin{figure}[t] 
    \centering
    \includegraphics[width=\columnwidth]{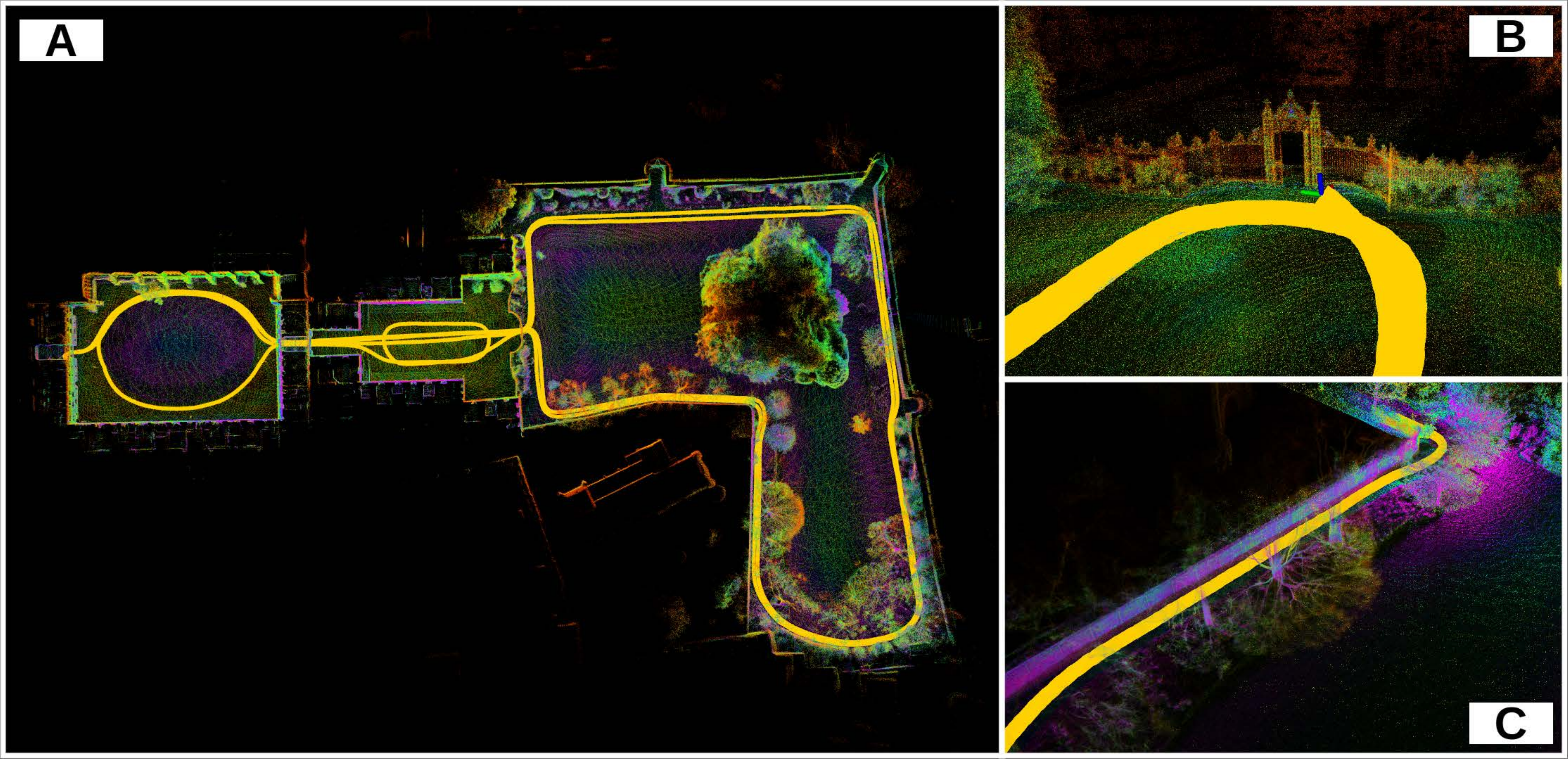}
    \caption{Accurate and detailed map produced by DFLIOM on Newer College Short (1.4 km) colored by intensity. Our method provides accurate trajectory (yellow) and map while significantly decreasing memory usage. Zoom-in showcase the details DFLIOM captures.}
    \label{fig:map}
    \vspace{-0.2in}
\end{figure}

The first main challenge is in developing algorithms that perform well for long duration missions.
For example, state-of-the-art direct approaches (register the full point cloud, not features) \cite{chen2023dlio, chen2023dliom} achieve high accuracy across many operating environments by proposing novel motion correction, keyframing, and loop closure modules.
However, using dense point clouds (albeit after a light voxelization filter) causes relatively expensive registration (memory), ultimately limiting the performance on long missions.
Alternatively, feature-based methods \cite{shan2020lio} extract edges and planar features using a series of hand-crafted heuristics \cite{zhang2017low, xu2021fast}.
Compared to direct methods, feature-based methods are more computationally efficient, but they tend to discard useful information, resulting in less robust performance.
In particular, these feature extractors do not consider the point cloud registration objective explicitly, and it is difficult to hand-craft good feature extraction heuristics solely based on the points' coordinates.
For example, the edge on a moving vehicle (transient) and the edge of a building (salient) appear identical to these feature extractors, thus relying solely on hand-crafted features can discard meaningful semantic context.

\begin{figure*}[t] 
    \centering
    \includegraphics[width=1.6\columnwidth]{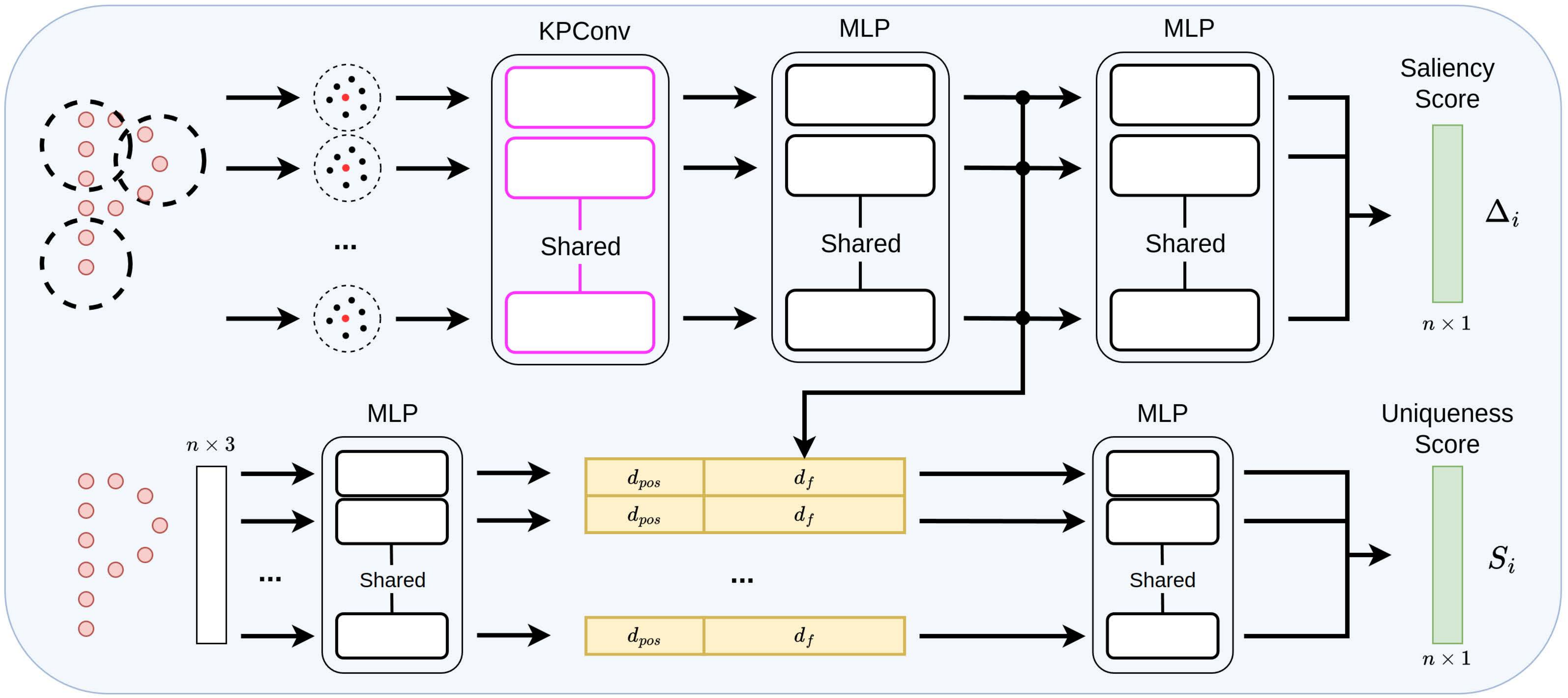}
    \caption{Architecture of the proposed feature extraction network. Kernel Point Convolution (KPConv) is used as the backbone for extracting a higher dimensional representation from point clusters. Inspired by PointNet we use shared MLPs to map point coordinates to higher dimension as positional embedding. Separate shared MLPs are used to predict the saliency and uniqueness scores, based on which more important points are chosen. Shared weights make our network sufficiently light-weight to inference in real-time.}
    \label{fig:architecture}
    \vspace{-0.2in}
\end{figure*}

Therefore, another challenge is to select a subset of points that are relevant to the SLAM or registration objective.
A promising recent approach in registration is to use learning-based keypoint detection algorithms \cite{li2019usip, bai2020d3feat, lu2020rskdd}, which can find accurate transformations while using only a small fraction of the full point cloud. 
Unlike classical feature-based methods, unsupervised or weakly supervised learning-based methods can explore a higher dimensional feature space (e.g., ignoring transient objects), which is a key to their resiliency in complex environments.
However, existing keypoint detectors designed for registration are not well-suited for SLAM.
Ref. \cite{bai2020d3feat} uses a U-Net with several KPConv layers, which is too computationally heavy for real-time inference, while real-time algorithms (e.g., \cite{lu2020rskdd}) typically generate ``superpoints'' (weighted average of points within a cluster), which can degrade registration accuracy between relatively distant scans.
Nonetheless, another advantage of learning to extract features is that the representations and backbones could be re-used in other learned components of SLAM systems (e.g., loop closure detection \cite{arce2023padloc} and place recognition \cite{ma2022overlaptransformer}).

To this end, this paper presents \textbf{D}eep \textbf{F}eature Assisted \textbf{L}iDAR \textbf{I}nertial \textbf{O}dometry and \textbf{M}apping (DFLIOM), an extension of DLIOM \cite{chen2023dliom},
with the following key contributions:
\begin{itemize}
    \item a light-weight learned feature extractor which selects points relevant for scan-to-scan and scan-to-map point cloud registration, enabling better localization accuracy with significantly less memory usage,
    \item an extension of a state-of-the-art LIO system (DLIOM) to leverage the proposed feature extractor, with an ability to fall back to dense mode to maintain robust performance across diverse environments,
    \item demonstrations that the feature extractor outperforms handcrafted ones, that the proposed SLAM system outperforms DLIO and DLIOM, two state-of-the-art LIO systems, in both localization accuracy and memory usage while running in real time, and that the approach generalizes to new locally collected datasets.
\end{itemize}

\noindent An example map produced by DFLIOM is shown in \cref{fig:map}.


\section{Preliminaries}

\subsection{Direct LiDAR Inertial Odometry and Mapping}

Direct LiDAR-Inertial Odometry and Mapping (DLIOM) \cite{chen2023dliom} is a state-of-the-art real-time SLAM algorithm based on accurate constant-jerk motion correction and components perceptive to the environment.
An incoming point cloud is first deskewed using a continuous-time motion correction module, and registered onto a submap built from selected keyframes using Generalized ICP \cite{segal2009generalized}.
Compared with previous work, DLIOM provides a convergence guarantee by employing a contracting geometric observer \cite{lopez2023contracting} for state estimation, which helps avoid inconsistent sensor fusion. 
Details of DLIOM's system architecture can be found in \cite{chen2023dliom}.

Although DLIOM achieves superior localization accuracy in real time, the use of dense point clouds contributes to memory consumption issues mentioned above, which limits its performance for longer missions.
Therefore, this paper builds upon DLIOM with the goal of achieving similar localization accuracy with significantly less memory usage.

\subsection{Kernel Point Convolution}

Kernel Point Convolution (KPConv) \cite{thomas2019kpconv} is an efficient extension of image convolution to point clouds. 
This operation is intrinsically invariant to translation and shown to be robust under varying point densities, and thus we use it as our network backbone for extracting a meaningful representation of point cloud neighborhood geometries.
Let $x_i$ denote a point from point cloud $\mathcal{P}$ and $f_i$ the corresponding feature from $\mathcal{F}$, and $\mathcal{N}_x$ the set of points within radius $r \in \mathbb{R}$ from $x$, i.e. $\mathcal{N}_x = \{ x_i \in \mathcal{P}: \|x_i-x\| \leq r \}$. 
Let $h$ denote linear correlation between kernel point $\hat{x}_k$ and point $x_i$ defined as

\begin{equation} \label{eq:correlation}
    h(x_i-x, \hat{x}_k) = \text{max}(0, 1- \frac{\|x_i-x - \hat{x}_k\|}{d_\sigma}) ,
\end{equation}
where $d_\sigma$ is the influence distance of the kernel points. Let $\mathbf{W}_k$ denote the weight matrix of $\hat{x}_k$, and $K$ the number of kernel points, the kernel function $g$ is defined as:
\begin{equation} \label{eq:kernel}
    g(x_i - x) = \sum_{k=1}^{K} h(x_i-x, \hat{x}_k) \mathbf{W}_k .
\end{equation}
And finally the point convolution of $\mathcal{F}$ by kernel $g$ at point $x \in \mathcal{P}$ is defined as:
\begin{equation} \label{eq:kpconv}
    (\mathcal{F} * g) (x) = \sum_{x_i \in \mathcal{N}_x} g (x_i - x) f_i .
\end{equation}

\section{Method}

Upon receiving a point cloud, we first account for motion distortion and apply a light voxelization filter following the procedure in \cite{chen2023dlio, chen2023dliom}.
Unlike previous work that extracts edge and planar features using handcrafted heuristics \cite{zhang2017low, shan2018lego, shan2020lio}, we propose to use a combination of \textit{salient} features (features that persist across scans) and \textit{unique} features (points useful for local-scale scan-to-scan matching).
We introduce our inference pipeline in \cref{sec:method:pipeline}.

\subsection{Salient Feature}

\begin{figure*}[t!] 
    \centering
    \begin{subfigure}[b]{0.3\linewidth}
    \includegraphics[width=\columnwidth]{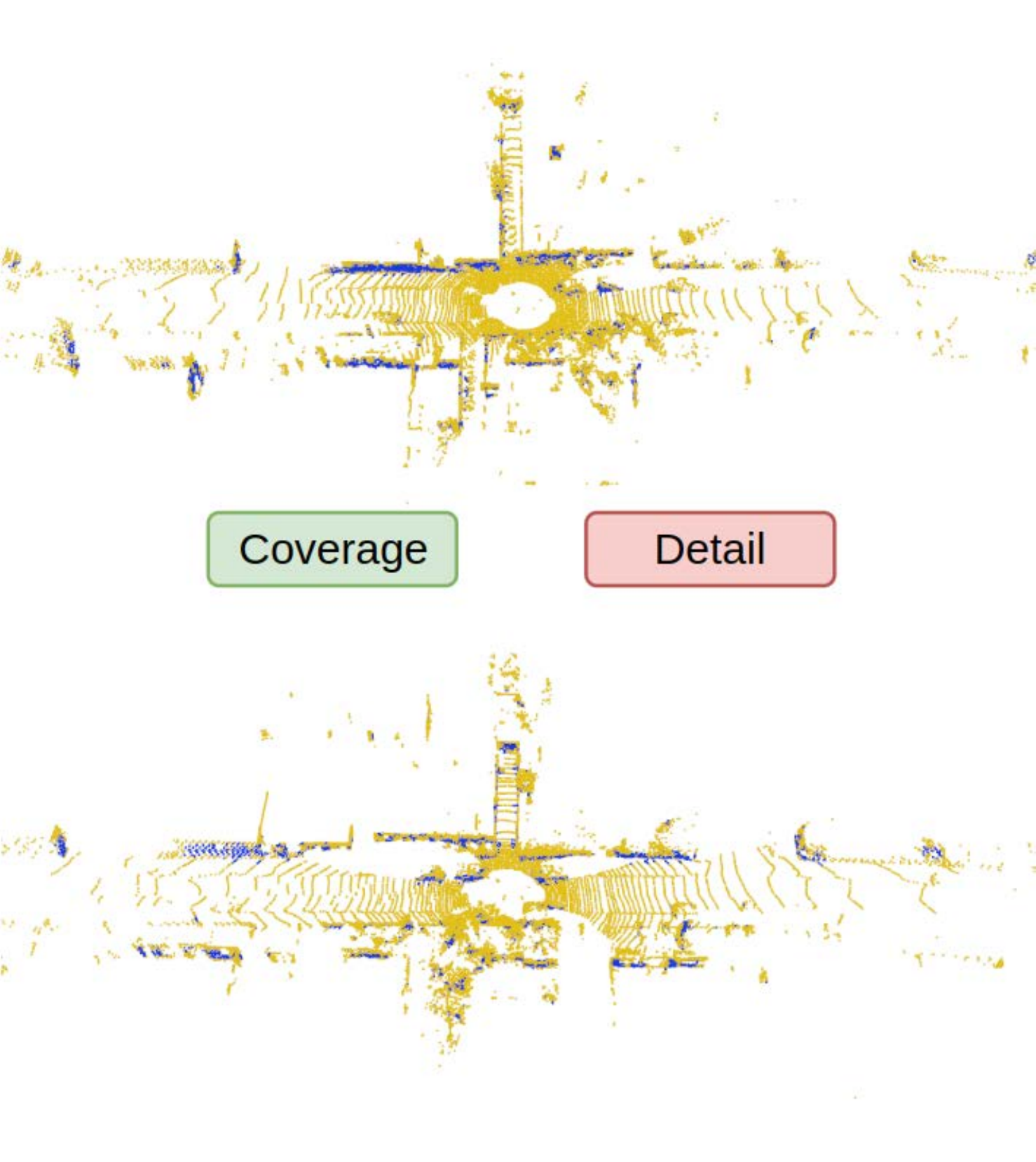}
    \caption{Only Saliency Features}
    \label{fig:diff:sal}
    \end{subfigure}%
    \begin{subfigure}[b]{0.3\linewidth}
    \includegraphics[width=\columnwidth]{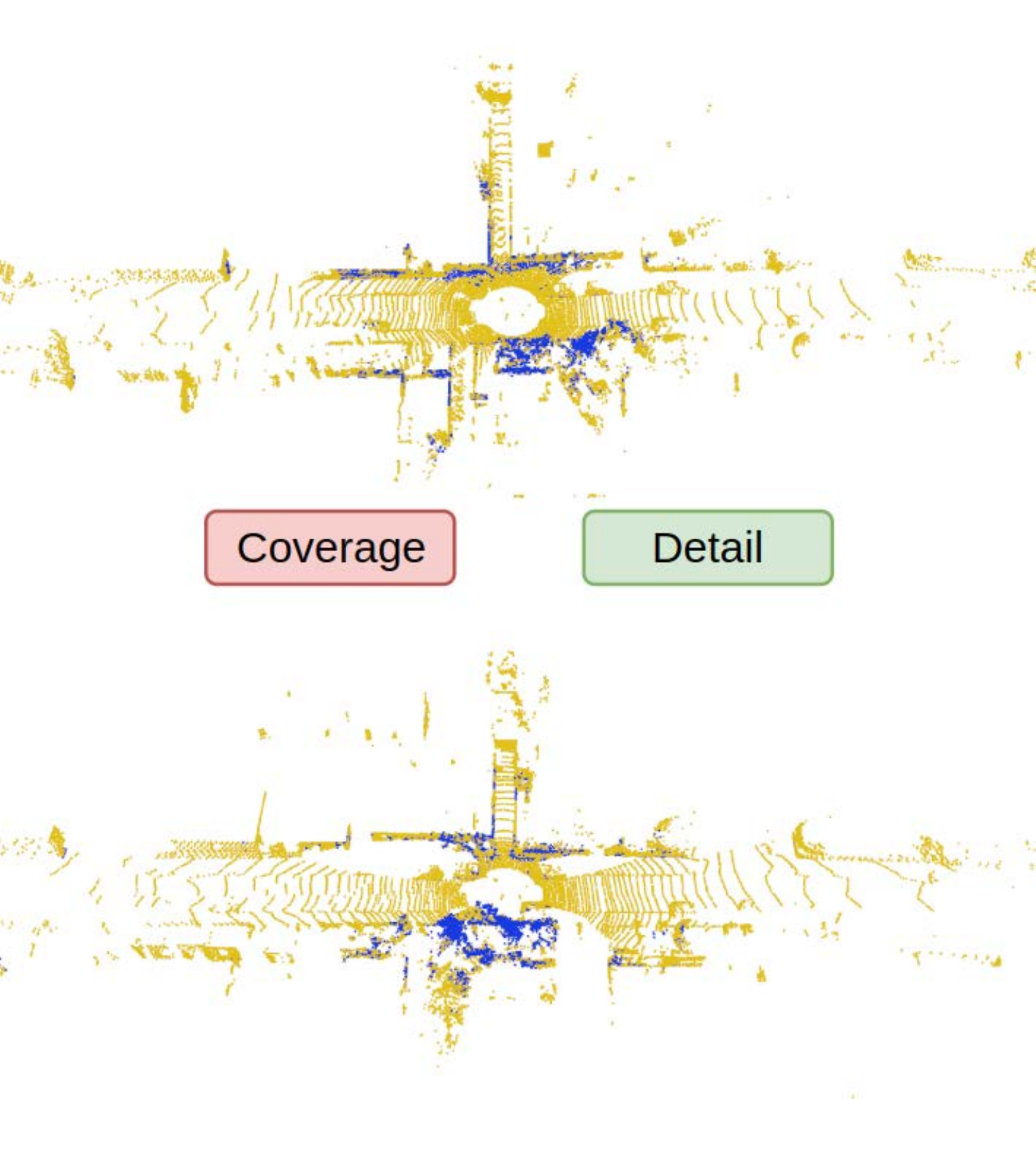}
    \caption{Only Unique Features}
    \label{fig:diff:uni}
    \end{subfigure}%
    \begin{subfigure}[b]{0.3\linewidth}
    \includegraphics[width=\columnwidth]{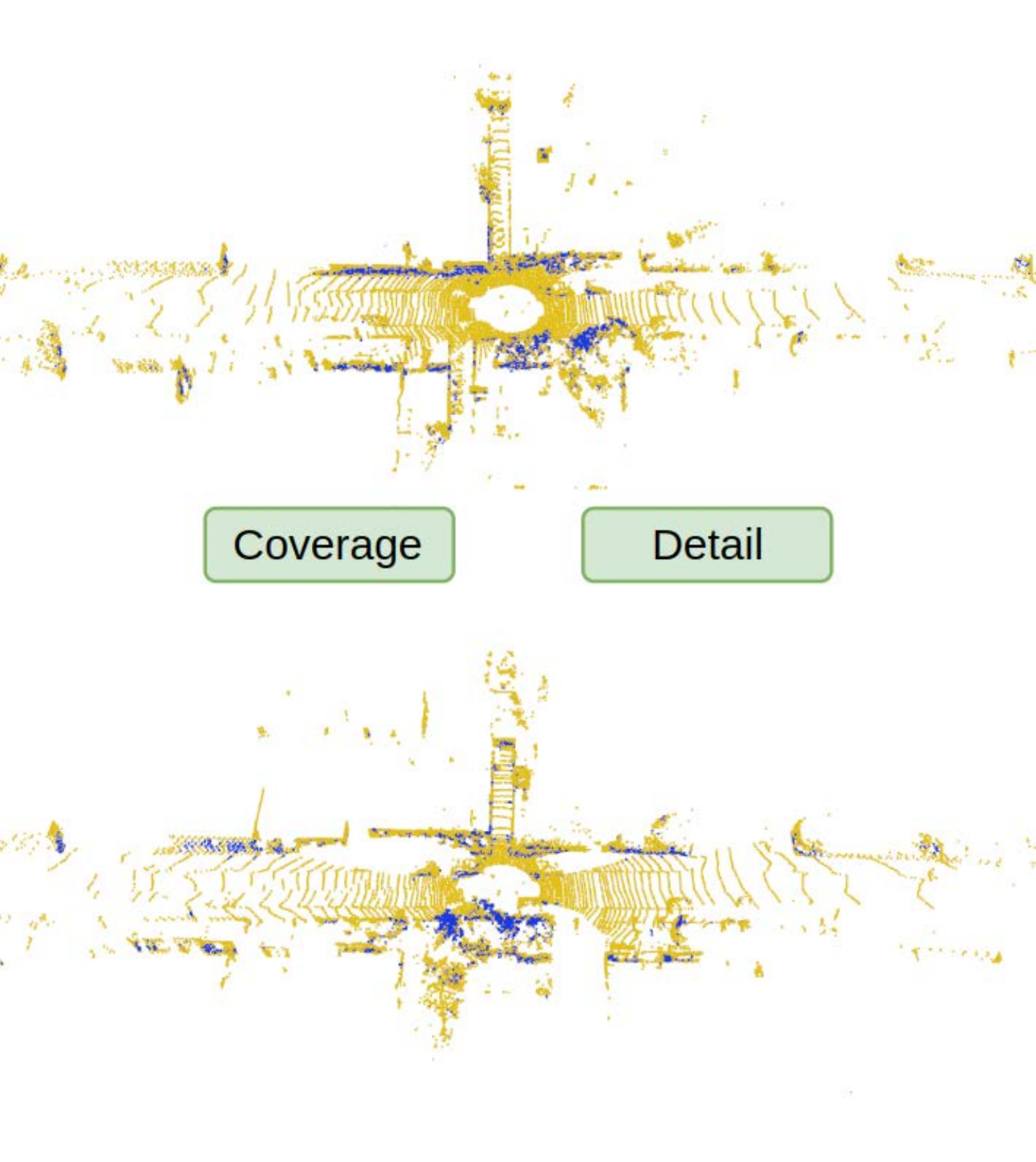}
    \caption{Both Types of Features}
    \label{fig:diff:both}
    \end{subfigure}
    \caption{Example point clouds after feature extraction. blue: selected feature points. (a): When only selecting best Saliency Features, parallel walls are selected, which can be featureless and similar to neighboring scans. (b): When only selecting best Unique Features, detailed features near the robot are selected and thus only useful for local scale scan-to-scan matching. (c): With both types of features, the extracted point clouds are rich in feature and provides good coverage.}
    \vspace{-0.15in}
\end{figure*} 


ICP-based algorithms heavily rely on finding good point correspondences to compute accurate relative transformations between two point clouds.
Therefore, when using these algorithms for SLAM with a moving sensor, it is crucial for the downsampled point clouds to contain points that persist across scans, which we denote as \textit{salient features} in this paper.
Unlike prior work that assumes useful points are edges and planes \cite{shan2020lio}, the proposed network, whose architecture is specified in \cref{fig:architecture}, learns to select such points directly.
It consists of a shared KPConv \cite{thomas2019kpconv} backbone and several shared MLPs, inspired by the success of PointNet \cite{qi2017pointnet}.

Given a dataset of point clouds $\mathcal{P}_0, \mathcal{P}_1, \cdots, \mathcal{P}_n$ and their ground truth poses $\mathbf{T}_0, \mathbf{T}_1, \cdots, \mathbf{T}_n$, we first obtain source and target point clouds $\mathcal{P}_i$ and $\mathcal{P}_{i-b}$ separated by $b$ scans.
The point clouds are first transformed onto a world frame using their ground truth pose provided in the dataset.
Points in $\mathcal{P}_i$ that have a neighbor in $\mathcal{P}_{i-b}$ within a distance $r$ are identified as $\mathcal{P}_i^+ \subseteq \mathcal{P}_i$, with the others denoted as $\mathcal{P}_i^- = \mathcal{P}_i \setminus \mathcal{P}_i^+$.
For every point $x_k \in \mathcal{P}_i$, point cluster $C_k$ is formed via radius search, and for every cluster $C_k$, we use the shared KPConv \cite{thomas2019kpconv} backbone to encode its geometric features.
A shared MLP is used to predict saliency score $\delta_k$, and the saliency scores associated with point cloud $\mathcal{P}_i$ is denoted $\mathbf{\Delta}_i$.
The network is trained using the hinge loss:
\begin{equation} \label{eq:saliency_loss}
    L_\text{sal} = \text{max}(0, \frac{\sum_{x_k \in \mathcal{P}_i^-} \delta_k}{|\mathcal{P}_i^-|} - \frac{\sum_{x_k \in \mathcal{P}_i^+} \delta_k}{|\mathcal{P}_i^+|} + M_\text{sal}),
\end{equation}
where $M_\text{sal}$ is the saliency loss margin, and $|\cdot|$ denotes set cardinality.
\cref{eq:saliency_loss} encourages the network to learn a significant margin ($M_\text{sal}$) between the score for points in $\mathcal{P}_i^+$ and points in $\mathcal{P}_i^-$, i.e. higher saliency scores for $x_k \in \mathcal{P}_s^+$.


However, if only points with high saliency scores are selected, the downsampled point clouds can exhibit high self-similarity, which causes poor registration results.
\cref{fig:diff:sal} shows two example point clouds when only selecting salient features (blue).
The selected feature points are along the two parallel walls on the sides of the road, and largely ignores the transient objects (cars parked, pedestrians, etc.) and ground points.
The GICP objective is to minimize the Mahalanobis distance between corresponding point pairs, and the covariances are structured to have small uncertainty along the surface normal and high uncertainty along the local plane.
The selected features in point clouds in \cref{fig:diff:sal} mainly consist of points along the same walls, and it can cause GICP to fall into a local minimum, resulting in slippage along the road (wall) direction.
To address this, we propose to also use \textit{unique features} to achieve better localization accuracy.

\subsection{Unique Feature}

We define \textit{unique features} as points that are useful for matching a scan to another scan(s) that is close to the robot's current position.
In LiDAR odometry algorithms, ICP \cite{besl1992method} and Generalized-ICP \cite{segal2009generalized} are two commonly used registration algorithms.
Existing learning-based point cloud registration algorithms \cite{lu2021hregnet, wang2019deep} essentially learn to predict higher scores for points useful for ICP by employing weighted SVD (differentiable) in the training loop for solving the ICP objective \cite{papadopoulo2000estimating}.
However, unlike ICP, GICP does not have a closed form solution, and thus it cannot be included in the training loop.
As a result, we propose a query-based strategy to obtain point-wise contribution to GICP. Pseudocode of the procedure is shown in \cref{alg:GICP_usefulness}.

\begin{algorithm}[H]
    \caption{GICP Usefulness Generation}
    \begin{algorithmic}[1] \label{alg:GICP_usefulness}
        \renewcommand{\algorithmicrequire}{\textbf{Input:}}
        \renewcommand{\algorithmicensure}{\textbf{Output:}}
        \REQUIRE $\{\mathcal{P}_0, \mathcal{P}_1, \cdots, \mathcal{P}_n\}, \{\mathbf{T}_0, \mathbf{T}_1, \cdots, \mathbf{T}_n\}, b, v$ \\
        \ENSURE $\{\mathcal{S}_b, \mathcal{S}_{b+1}, \cdots, \mathcal{S}_{n-b}\}$ \\
        \FOR{$i \in [b, n-b]$} 
            \STATE $\mathcal{P}_s \leftarrow \mathbf{T}_i \mathcal{P}_i; \quad \mathcal{P}_t \leftarrow \mathcal{P}_{i-b} \cup \mathcal{P}_{i+b}$ \\
            \STATE $\Tilde{\mathcal{P}_s}, \Tilde{\mathcal{P}_t} \leftarrow \text{voxelize}(\mathcal{P}_s, v), \text{voxelize}(\mathcal{P}_t, v)$ \\
            \STATE $\{\mathcal{M}_0, \mathcal{M}_1, \cdots, \mathcal{M}_k\} \leftarrow \text{preprocess}(\mathcal{P}_s)$ \\
            \FOR{$j \in [k]$}
                \STATE $\hat{\mathcal{P}}_s \leftarrow \Tilde{\mathcal{P}_s} \cup \mathcal{P}_s[\mathcal{M}_j]$ \\
                \STATE $\hat{\mathcal{P}}_t \leftarrow \Tilde{\mathcal{P}_t} \cup \text{getNeighbors}(\mathcal{P}_t, \mathcal{P}_s[\mathcal{M}_j])$ \\
                \STATE $\hat{\mathbf{T}}_i \leftarrow \text{GICP}(\hat{\mathcal{P}}_s, \hat{\mathcal{P}}_t)$ \\
                \STATE $\mathbf{S}_i[\mathcal{M}_j] \leftarrow \frac{\text{Err}(\hat{\mathbf{T}}_i, \mathbf{T}_i)}{|\mathcal{M}_j|}$
            \ENDFOR
        \ENDFOR
    \end{algorithmic}
\end{algorithm}

Let point cloud $\mathcal{P}_i$, transformed by (known) random transformation $\mathbf{T}_i$, be the source point cloud, and $\mathcal{P}_{i-b} \cup \mathcal{P}_{i+b}$ the target point cloud.
$\mathcal{P}_s$ and $\mathcal{P}_t$ are first voxelized to obtain skeleton point clouds $\Tilde{\mathcal{P}_s}$ and $\Tilde{\mathcal{P}_t}$.
We run ground removal on $\mathcal{P}_s$ and divide the rest into patches $\{\mathcal{M}_0, \mathcal{M}_1, \cdots, \mathcal{M}_k\}$.
For each patch $\mathcal{M}_j$, we add it to the source skeleton $\Tilde{\mathcal{P}_s}$ and its closest neighbors in $\mathcal{P}_t$ to $\Tilde{\mathcal{P}_t}$, and denote the resulting point clouds $\hat{\mathcal{P}}_s$ and $\hat{\mathcal{P}}_t$.
We estimate the relative transformation $\hat{\mathbf{T}}_i$ using GICP, and record the error between $\hat{\mathbf{T}}_i$ and $\mathbf{T}_i$, normalized by the number of points in the patch, i.e. the cardinality of $\mathcal{M}_j$, to obtain uniqueness score, denoted $\mathbf{S}_i$.
The process is repeated for all point clouds in the dataset, and the lower the error recorded, the more useful the point is to GICP.
During training, the network is trained to predict, for uniqueness score $s_k, \forall x_k \in \mathcal{P}_i$ using the MSE loss (ground truth obtained via \cref{alg:GICP_usefulness}).


\cref{fig:diff:uni} shows the point clouds when only selecting the highest-scoring unique features.
Unlike the parallel walls, this type of feature is unique at a local scale, and thus helps in figuring out where the robot is along the road, significantly improving performance in repetitive environments.
However, SLAM involves scan matching beyond a local scale.
As a result, both types of features should be utilized in feature extraction.
As shown in \cref{fig:diff:both}, when using both salient and unique features, the point clouds differ from each other significantly, while also providing good scan coverage for scan-to-map registration.
The effect of adding unique features is further demonstrated in \cref{sec:exp:ablation} through ablation study.

\subsection{Inference Pipeline}
\label{sec:method:pipeline}


Upon receiving point cloud $\mathcal{P}_i$, we first motion deskew the point cloud using recent IMU measurements $\alpha_k, \omega_k$ as described in \cite{chen2023dlio, chen2023dliom}, and voxel downsample the deskewed point cloud to obtain processed point cloud $\Tilde{\mathcal{P}}_i$.
When the robot is in a small-scale environment (e.g. narrow corridor), the voxelized point clouds contain fewer points compared to when the robot is in an open environment.
Further downsampling point clouds in such environments can result in important details being dropped, thus negatively impacting localization accuracy.
Let $D_i$ denote the average Euclidean distance between the robot and the points, we perform feature extraction on point cloud $\Tilde{\mathcal{P}}_i$ if $D_i > d$ where $d$ is a predefined hyper-parameter.
Switching between dense and feature-based modes allows our method to minimize memory usage while maintain robust performance in environments where existing feature-based methods would easily fail.

During feature extraction for point cloud $\Tilde{\mathcal{P}}_i$, we first predict the saliency scores $\mathbf{\Delta}_i$ and uniqueness scores $\mathbf{S}_i$. 
We select the highest-scoring salient and unique features, and combine them with a skeleton point cloud obtained via aggressive voxel filter with grid size $v_s$.
When many points are close to $R$ away from the robot where $R$ is the sensing range of the LiDAR, the robot is likely in a wide open field or the LiDAR range cannot fully cover the environment, and important points might be dropped because of the extremely low point density in that region.
As a result, when there are more than $m$ points $\alpha R$ away from the robot, we keep those points in the downsampled point cloud as well, and denote them \textit{coverage features}.
The downsampled point cloud $\mathcal{P}_i^*$ is used for downstream processing (e.g. registration, loop closure, and submapping).

Selecting proper keyframes for submap is crucial for accurate localization.
DLIOM \cite{chen2023dliom} selects keyframes using 3D Jaccard index (a type of IoU).
This metric, although perceptive to the environment, is computationally expensive and sensitive to varying point density in the point cloud after feature extraction, leading to constant rebuilding of the submaps and thus unnecessary memory allocations.
As a result, similar to \cite{chen2023dlio}, we associate every keyframe with the robot's state when the keyframe is placed, and use the $K$ nearest keyframes and $L$ nearest convex hull keyframes as the submap.
Compared with the 3D Jaccard index-based submapping, this approach is light-weight and invariant to point density, avoiding repeatedly rebuilding submaps. 
This inference pipeline is summarized in \cref{alg:pipe}.

\begin{algorithm}[h]
    \caption{DFLIOM}
    \begin{algorithmic}[1] \label{alg:pipe}
        \renewcommand{\algorithmicrequire}{\textbf{Input:}}
        \renewcommand{\algorithmicensure}{\textbf{Output:}}
        \REQUIRE $\mathcal{P}_i, \alpha_k, \omega_k, \hat{\mathbf{X}}_{i-1}, d, v_s, m, R, \eta, K, L$ \\
        \ENSURE $ \hat{\mathbf{X}}_{i}$ \\
        \STATE $\Tilde{\mathcal{P}}_i, D_i \gets \text{preprocess}(\mathcal{P}_i, \alpha_k, \omega_k, \hat{\mathbf{X}}_{i-1})$ \\
        \IF{$D_i \leq d$}
            \STATE $\mathcal{P}_i^* \gets \Tilde{\mathcal{P}}_i$ \\
        \ELSE
            \STATE $\mathbf{\Delta}_i, \mathbf{S}_i \gets \text{modelInference}(\Tilde{\mathcal{P}}_i)$ \\
            \STATE $\mathcal{P}_i^* \gets \text{voxelize}(\Tilde{\mathcal{P}}_i) \cup \text{selectBestFeatures}(\Tilde{\mathcal{P}}_i, \mathbf{\Delta}_i, \mathbf{S}_i)$ \\
            \IF{$|\{x \in \Tilde{\mathcal{P}}_i : \|x\| \geq \eta R \}| \geq m $}
                \STATE $\mathcal{P}_i^* \gets \mathcal{P}_i^* \cup \{x \in \Tilde{\mathcal{P}}_i : \|x\| \geq \eta R \} $ \\
            \ENDIF
        \ENDIF
        \STATE $\hat{\mathbf{T}}_i \gets \text{GICP}(\mathcal{P}_i^*, \text{getSubmap}(K, L))$ \\
        \STATE Downstream Processing (e.g. Registration, Submapping, Loop Closure) \\
        \STATE return $\hat{\mathbf{X}}_i$
    \end{algorithmic}
\end{algorithm}

\section{Experiments}

\subsection{Experiment Setting}

Our feature extractor is implemented using PyTorch \cite{paszke2019pytorch}, and the rest is implemented in C++. 
Our experiments are conducted on a PC with an Intel i9-13900K CPU, Nvidia RTX 4090 GPU, and 64 GB of RAM.
Due to the lack of ground truth semantic labels in Newer College Dataset \cite{ramezani2020newer}, we first train the backbone and saliency head using a subset of Newer College Dataset Short (with saliency loss margin $M_{sal} = 0.5$ and scan bias $b = 150$), freeze the trained layers, prepare GICP usefulness scores using 3000 scan pairs from Semantic KITTI sequence 00 \cite{behley2019iccv} (with voxel size $v = 1.5$ and scan bias $b = 10$), and train the uniqueness head for 200 epochs. 
For coverage features, we set $\alpha = 0.9$ and $m = 0.01 |\mathcal{P}_i|$. 
During preprocessing, we set the voxel filter grid size to 0.25 meters. In submapping we set $K$ and $L$ such that at most 20 keyframes are selected to make sure we use same number of keyframes as DLIOM \cite{chen2023dliom}.
In experiments, we found selecting best 10\% salient and 10\% unique features suffices for achieving accurate localization on the tested sequences.
All tables report the mean of 3 runs.
RAM usage, according to our experiments, may vary drastically across different hardware and system setup.

\subsection{Newer College Dataset}

\begin{table*}
    \centering
        \begin{tabular}{c | c | c c c c | c c c c | c}
        \hline
        \multicolumn{2}{c|}{ } & \multicolumn{4}{|c|}{Newer College \cite{ramezani2020newer}} & \multicolumn{4}{|c|}{Newer College Extension \cite{zhang2021multicamera}} & \\
        \hline
        Metric & Algorithm & Short & Mound & Long & Quad w. Dyn. & Math Hard & Quad Hard & Cloister & Park & Avg. \\
        \hline
        \multirow{3}{4em}{RMSE [m] $\downarrow$ }
        & DLIOM \cite{chen2023dliom} & 0.452          & 0.200          & 0.450          & 0.155          & \textbf{0.067} & 0.066       & 0.093       & \textbf{0.304} & + 0.0 \% \\
        & DLIO \cite{chen2023dlio}   & 0.463          & 0.210          & 0.482          & 0.172          & 0.110          & 0.108       & 0.133       & 0.334       &  + 24.3 \% \\
        & Ours                       & \textbf{0.409} & \textbf{0.188} & \textbf{0.398} & \textbf{0.147} & 0.077       & \textbf{0.065} & \textbf{0.092} & 0.342       & \red{- 2.4 \%}  \\
        \hline
        \multirow{3}{4em}{RAM [GB] $\downarrow$ }
        & DLIOM & 14.61          & 13.00          & 14.02         & 12.13          & 5.60          & 5.37          & 10.56          & 14.32         & + 0.0 \% \\
        & DLIO  & 11.20          & 10.56          & 13.58         & 4.39           & 6.67          & 5.66          & 8.88           & 13.70         & -4.2 \% \\
        & Ours  & \textbf{5.45}  & \textbf{3.99}  & \textbf{6.59} & \textbf{1.71}  & \textbf{2.77} & \textbf{2.21} & \textbf{5.19}  & \textbf{8.52} & \red{- 57.5 \%}  \\
        \hline
        \multirow{3}{4em}{Runtime [ms] $\downarrow$ }
        & DLIOM & \textbf{12.17} & 24.74            & 14.17            & 24.19            & 28.56            & 28.12            & 23.92           & 28.81        & 25.39 ms \\
        & DLIO  & 14.44          & \textbf{18.58}   & \textbf{13.85}   & \textbf{23.24}   & \textbf{23.12}   & \textbf{21.59}   & \textbf{23.11}  & \textbf{26.55}  & \red{22.44 ms} \\
        & Ours  & 36.64          & 35.67            & 39.52            & 37.07            & 42.12            & 35.58            & 24.30           & 47.03        & 40.03 ms \\
        \hline
        \end{tabular}
    \caption{RMSE, Resident Set Size (RSS), and Runtime comparison to the baselines. We achieve lower RMSE using significantly less memory, while maintaining realtime performance (40.0 ms average runtime per scan). LIO-SAM \cite{shan2020lio} is not included as it does not support 6-axis IMU used in the dataset. Best results are marked with \textbf{Bold} or \red{Red}.}
    \label{Tab:ncl_results}
    \vspace{-0.15in}
\end{table*}

We first test DFLIOM on Newer College Dataset \cite{ramezani2020newer} (10 Hz OS1-64 LiDAR and the 100 Hz internal IMU), and its extension \cite{zhang2021multicamera} (10 Hz OS0-128 LiDAR and 200 Hz Alphasense Core IMU).
The downsampling threshold is set to $d = 3.0$, except in the Mound sequence $d = 5.0$ because of its more aggressive movements in a narrow environment.
In \cref{Tab:ncl_results} we show results of DFLIOM compared to the baselines DLIO \cite{chen2023dlio} and DLIOM \cite{chen2023dliom} on absolute trajectory error (RMSE), RAM usage (resident set size, RSS), and average processing time per frame.
Note that we select representative sequences in \cref{Tab:ncl_results} for space reasons, but we used all sequences in the datasets when computing the average statistics (except for the stairs sequence, as none of its scans satisfy $D_i \geq 3.0$).

\begin{figure}[t!] 
    \centering
    \begin{subfigure}[b]{0.5\linewidth}
    \includegraphics[width=0.95\columnwidth]{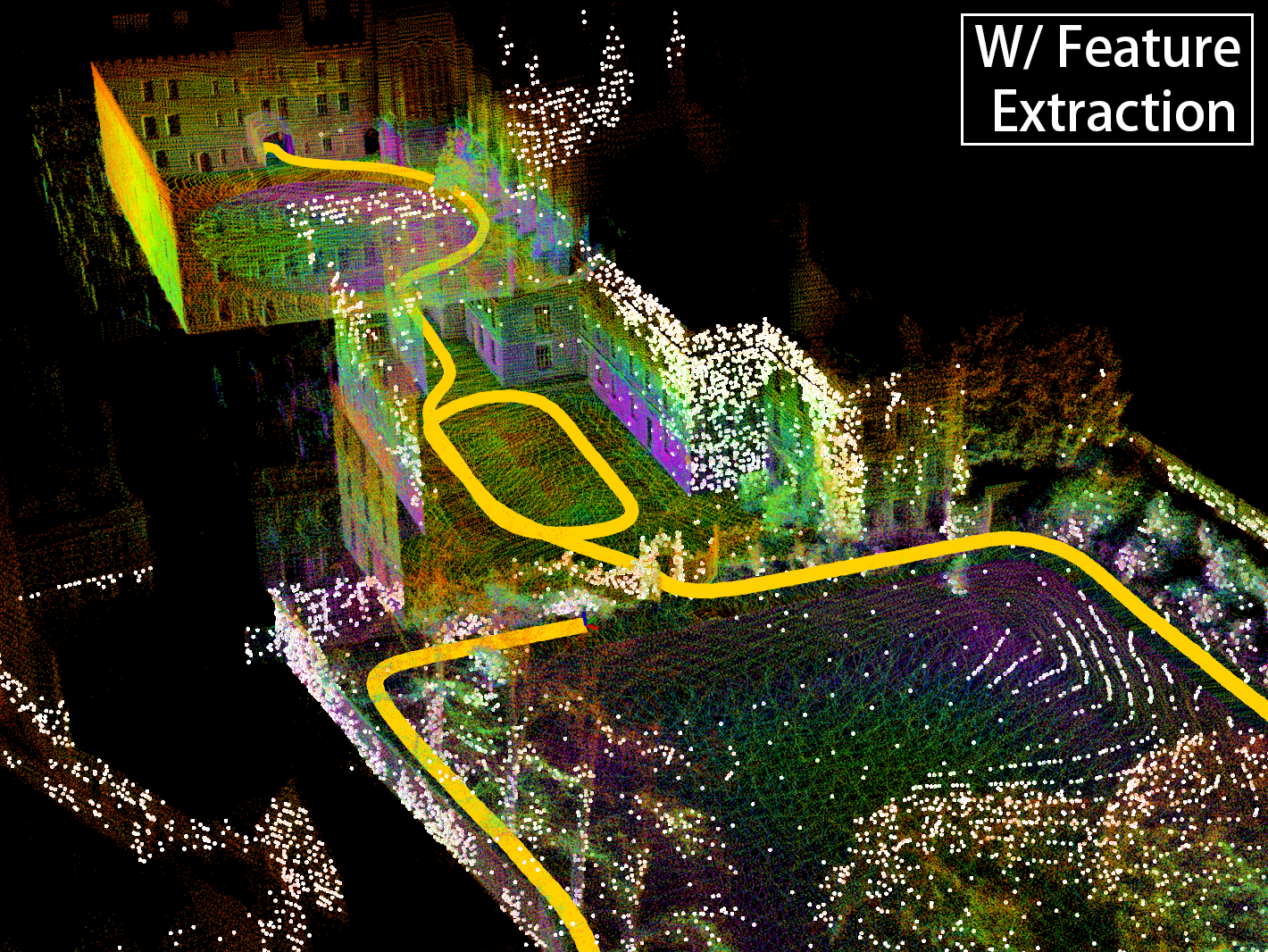}
    \end{subfigure}%
    \begin{subfigure}[b]{0.5\linewidth}
    \includegraphics[width=0.95\columnwidth]{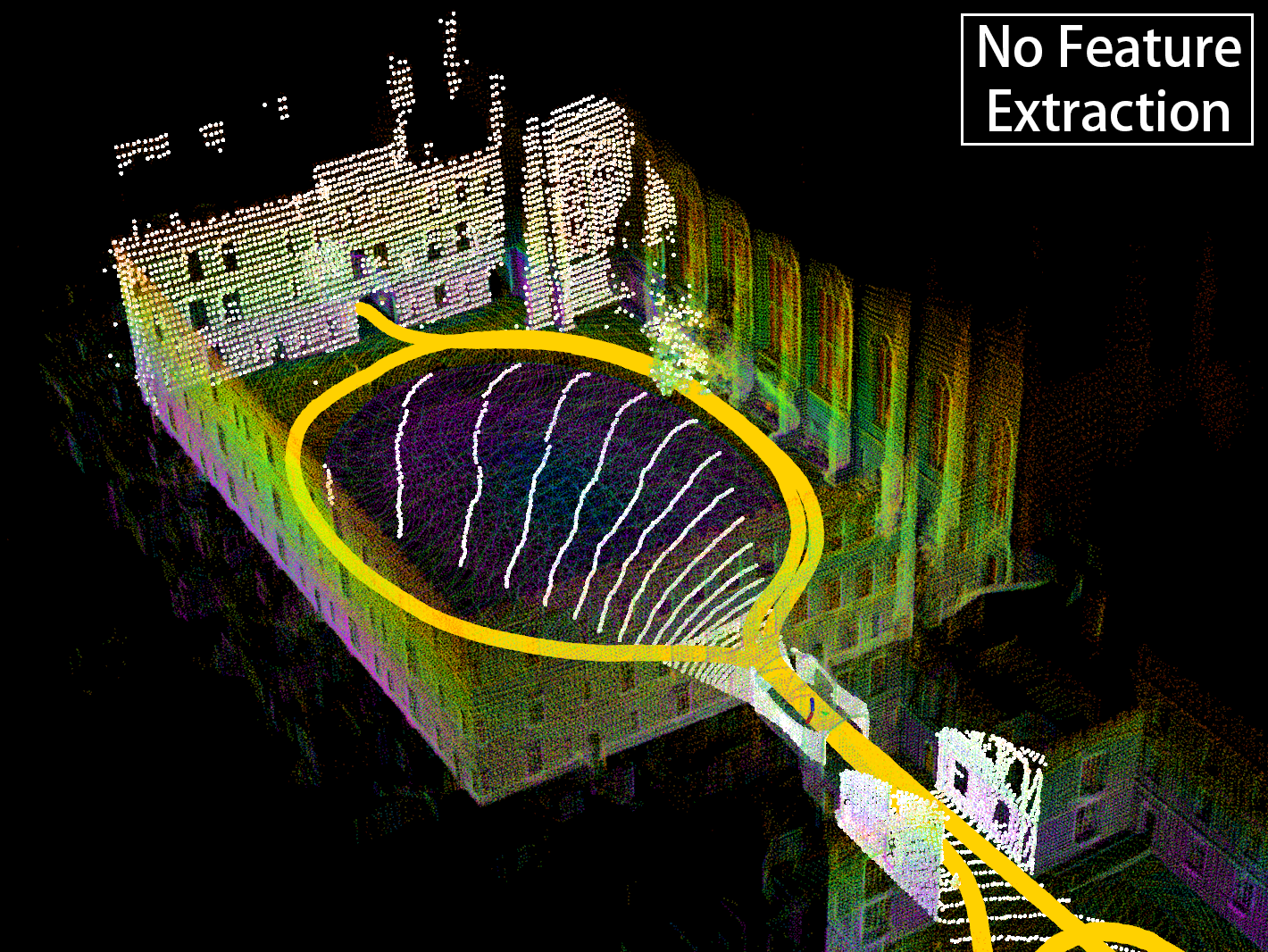}
    \end{subfigure}
    \caption{Example point clouds (colored by intensity) used by our method. (a): Point cloud after feature extraction (white) (b): Feature extraction is not performed when the robot is in a narrow corridor.}
    \label{fig:scan}
    \vspace{-0.2in}
\end{figure} 

\begin{figure}[t!] 
    \centering
    \includegraphics[width=0.9\columnwidth]{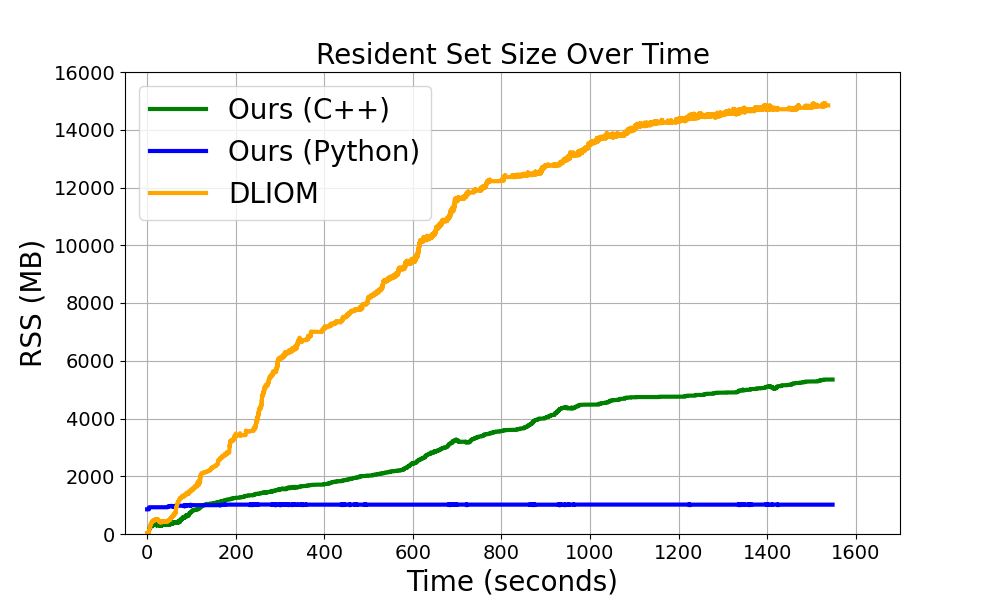}
    \caption{Memory usage (RSS in MB) vs. time along trajectory on Newer College Dataset \cite{ramezani2020newer} Short with DFLIOM (green) and DLIOM (orange). DFLIOM uses significantly less memory (e.g., from 14.6GB to 5.5GB) while maintaining localization accuracy. Our Python feature extractor (blue) uses constant RAM ($\sim$1GB).}
    \label{fig:mem_usage}
    \vspace{-0.2in}
\end{figure}

DFLIOM achieves better RMSE on most sequences and comparable on the rest, resulting in a 2.4\% RMSE reduction compared to vanilla DLIOM. 
In \cref{fig:scan}(a) we show an example point cloud (white) after our feature extraction module.
Although not explicitly designed for the purpose, our feature extractor largely ignores the transient objects, attending only to walls, tree trunks, a ramp across the parkland, and a detailed gate next to the robot.
This is a possible explanation for the gain in accuracy we achieved despite using significantly less points.
When the robot is in a narrow corridor, shown in \cref{fig:scan}(b), we do not perform feature extraction and use the dense point cloud directly to improve robustness in such regions.
As we only use roughly 20\% of the points in most cases, the memory and computation needed for processing deskewed point clouds (one of the major sources of memory allocation) is significantly reduced, achieving an overall RAM usage reduction of 57.5\% compared with DLIOM.
\cref{fig:mem_usage} shows the DKLIOM and DLIOM's resident set size vs. time on Newer College Dataset \cite{ramezani2020newer} Short, where DFLIOM (green) uses 5.5 GB and DLIOM (orange) uses 14.6 GB.
After the robot returns to a previously seen region near 800s (only few keyframes are placed during the rest of the trajectory), both methods' memory usage growth rate decrease.
At 800s, DFLIOM and DLIOM use 3.57 GB and 12.22 GB, respectively.
Assuming a constant growth of memory usage, we thus anticipate that our method is able to handle missions 3+ times as long as DLIOM can handle using the same amount of memory, i.e. with the same hardware RAM setup, while maintaining comparable or even better localization accuracy.
Our feature extraction network has 57K parameters, and uses shared layers for all point clusters, leading to an average runtime of 40.0 ms per frame.
Although we took a slight penalty on runtime, the ability to run in real-time with significantly less memory is a worthwhile trade-off for many autonomous systems.
Finally, our feature extractor is implemented in Python, and uses roughly constant RAM (1GB) and GPU memory (9.51GB), but these numbers can be reduced substantially with future memory optimization (e.g. C++ implementation).


\subsection{Comparison with LIO-SAM}

As LIO-SAM \cite{shan2020lio}, a feature-based baseline, does not support 6-axes IMU, we compare DFLIOM to LIO-SAM \cite{shan2020lio} on the campus loop in \cite{shan2020lio} (10 Hz Velodyne VLP16 LiDAR with 1000 Hz MicroStrain 3DM-GX5-25 IMU).
Due to the lack of ground truth, we compare end to end translation error as in \cite{shan2020lio}.
To better reflect drift along the trajectory, we disable loop closures on all tested methods.
For fair comparison, GPS in LIO-SAM is also disabled and we increase its CPU cores from 4 to 32 for faster runtime.
DFLIOM, even without benefiting from the three additional IMU axes, achieves lower End to End Translation Error and faster runtime than LIO-SAM.
This experiment further demonstrates our feature extractor's generalizability to much sparser LiDARs.

\begin{table}[t]
    \centering
        \begin{tabular}{c | c c }
        \hline
         & Trans. Error [m] & Runtime [ms] \\
        \hline
        LIO-SAM \cite{shan2020lio}  &  12.28    & 54.00    \\
        Ours                        &  11.20    & 31.61    \\
        \hline
        \end{tabular}
    \caption{End to End Translation Error and Runtime comparison with LIO-SAM on LIO-SAM Campus Large dataset. Our method outperforms LIO-SAM on localization accuracy and runtime.}
    \label{Tab:lio_sam}
    \vspace{-0.15in}
\end{table}

\subsection{Northeastern Campus Dataset}
\begin{figure}[t] 
    \centering
    \begin{subfigure}[b]{0.48\linewidth}
    \includegraphics[width=0.96\columnwidth]{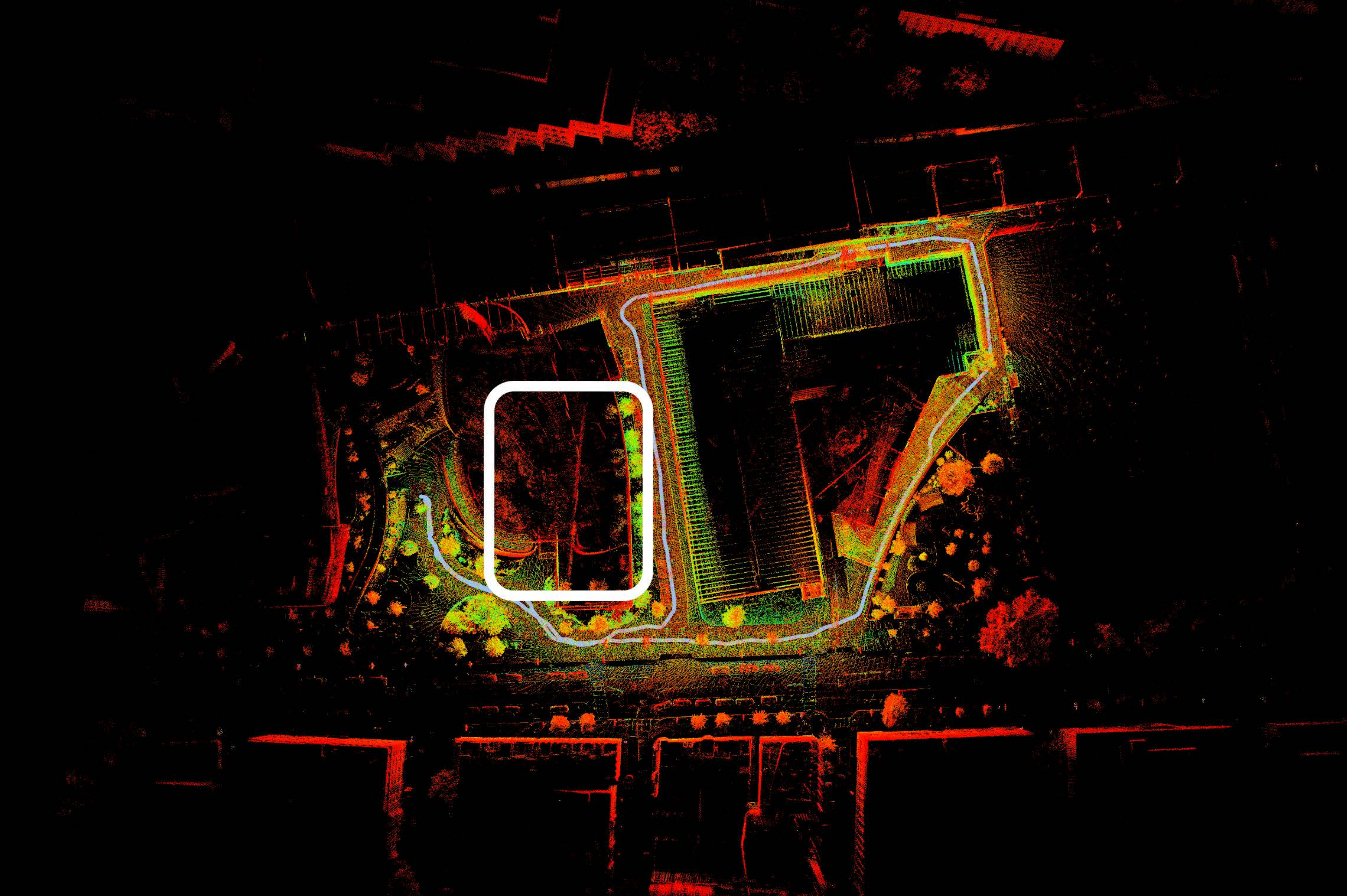}
    \label{fig:neu:exp_map}
    \end{subfigure}%
    \begin{subfigure}[b]{0.48\linewidth}
    \includegraphics[width=0.96\columnwidth]{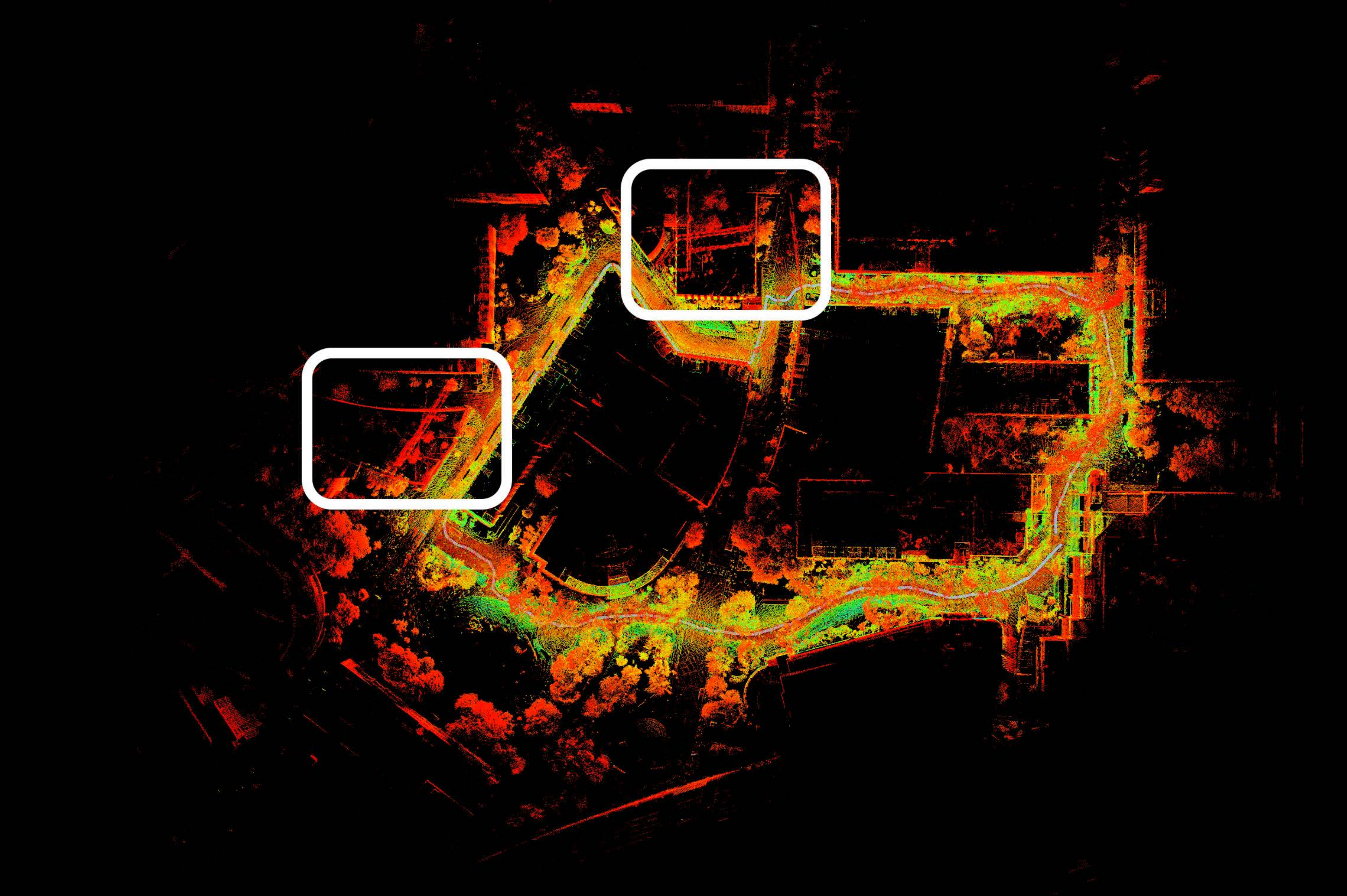}
    \label{fig:neu:campus_map}
    \end{subfigure}

    \vspace{0.02in}
    \begin{subfigure}[b]{0.48\linewidth}
    \includegraphics[width=0.96\columnwidth]{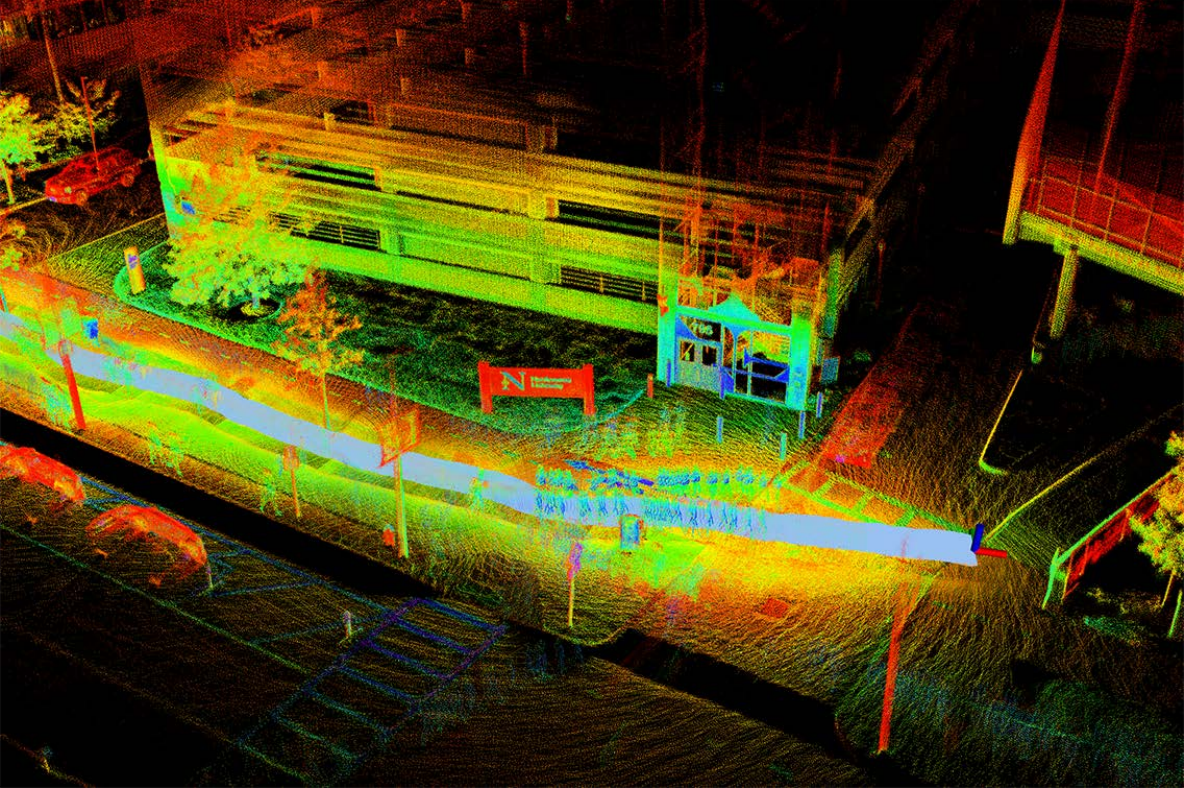}
    \label{fig:neu:exp_zoom}
    \end{subfigure}%
    \begin{subfigure}[b]{0.48\linewidth}
    \includegraphics[width=0.96\columnwidth]{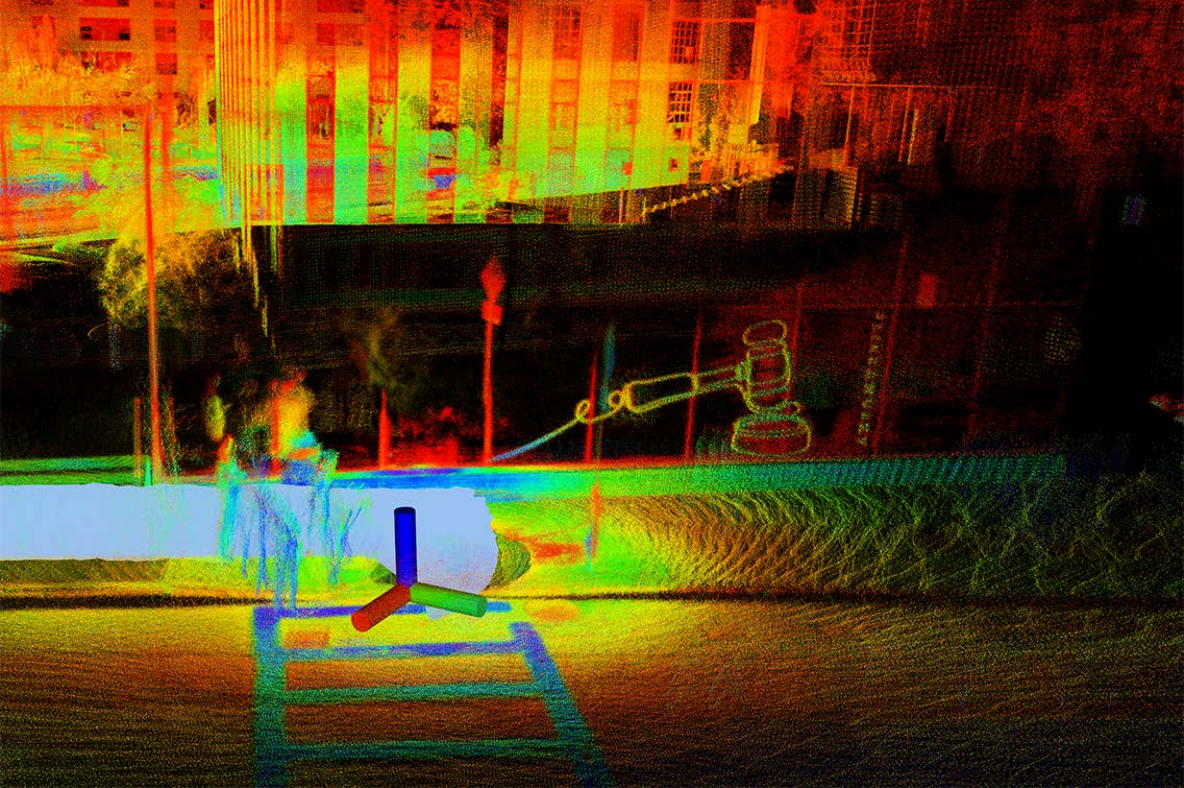}
    \label{fig:neu:campus_zoom}
    \end{subfigure}
    \caption{Keyframe Maps and Dense Reconstructions (all point clouds in the previous 100s overlaid) around Northeastern campus colored by intensity. Text and patterns in the environment are clearly visible.}
    \label{fig:neu}
    \vspace{-0.2in}
\end{figure}

To demonstrate DFLIOM's generalizibility to other environments, we ran our method on a self-collected dataset around our campus.
The dataset consist of two sequences, ISEC (548.32 m) and Main Campus (727.50 m), and is recorded with a 10 Hz OS1-128 LiDAR with its internal 100 Hz IMU.
The dataset is collected during a typical school day to maximize transient objects in the point clouds.
Since ground truth pose is not available, we show the reconstructed map, detailed in \cref{fig:neu}, as a proxy.
The Northeastern sign, parked vehicles, and LED light on the window are clearly visible, showcasing the level of accuracy our method is able to achieve despite the large amount of pedestrians and vehicles present.
Although the trajectory  passes several buildings with glass surfaces, the localization accuracy does not seem to be impacted by the reflections (white in \cref{fig:neu}).

\subsection{Ablation Study}
\label{sec:exp:ablation}

To further demonstrate the effectiveness of our learning-based feature extractor, we replaced our feature extraction module with various handcrafted ones and compared the performances.
For fair comparison, we try to select same number of points from all feature extractors tested.
For the random feature extractor, we simply select 20\% points randomly.
For salient/unique features only, we ignore the other prediction head to only train the network to extract salient or unique features for better performance.
LIO-SAM \cite{shan2020lio} utilizes a roughness-based handcrafted feature extractor, where points are classified into edge and surface features based on a predefined threshold.
To maintain a fair comparison, we selected the top 5\% points with highest roughness, and either randomly sample 15\% surface features (L-Rand) or voxelize the surface features with a grid size of 1.2 meters (L-Voxel) as it roughly selects 15\% points.

As shown in \cref{Tab:ablation}, our feature extraction module substantially outperform the others, including the feature extractor used by most previous feature-based methods \cite{zhang2017low, shan2018lego, shan2020lio}, on localization accuracy.
Notably, using only salient or unique features resulted in worse localization accuracy.
Moreover, as we do not need information like points' ring number and number of beams, our feature extractor generalizes to different LiDARs more easily, as demonstrated by DFLIOM's robust performance across datasets collected using different sensors.
\begin{table}[h]
    \centering
        \begin{tabular}{c | c c c c c c}
        \hline
         & Rand & Salient & Unique & L-Rand & L-Voxel & Ours\\
        \hline
        RMSE       & +7.9        & +2.4    & +14.9   & +9.2    & +6.4    & \red{-2.4} \\
        Runtime    & \red{27.8}  & 43.8    & 46.0    & 59.4    & 57.8    & 40.0       \\
        \hline
        \end{tabular}
    \caption{RMSE and Runtime change w.r.t. DLIOM \cite{chen2023dliom} comparison when replacing our feature extractor with hand-crafted one, including that of LIO-SAM. DFLIOM outperform the rest by a large margin in localization accuracy.}
    \label{Tab:ablation}
    \vspace{-0.2in}
\end{table}

\section{Conclusion}

In this paper, we present DFLIOM, a LIO system based on learned feature extractor to extract salient and unique features.
Our method improves localization accuracy and uses significantly less memory compared to DLIO and DLIOM, two state-of-the-art LIO systems, while running in real-time for 20 Hz LiDAR. 
Performance of DFLIOM is demonstrated on multiple public benchmarks collected using different LiDARs, and a self-collected dataset on our campus.
We further provide ablation study to showcase the effectiveness of the proposed feature extractor.
Future work will explore additional methods for incorporating registration in the training process.



\printbibliography

\end{document}